\documentclass[12pt]{article}
\usepackage{textcomp}
\usepackage[T1]{fontenc}
\usepackage[pdftex]{graphicx}
\usepackage{setspace}
\usepackage{multirow}
\usepackage{url}

\newcommand{\vsn}{\vspace{-0.35cm}\noindent}
\newcommand{\score}{\textit{score}}

\def\SeizOport{\mbox{\textit{SeizOport}}}
\def\SuccOper{\mbox{\textit{SuccOper}}}
\def\AvgRetOper{\mbox{\textit{AvgRetOper}}}
\def\Score{\mbox{\textit{Score}}}

\newcommand{\qed}{\nobreak \ifvmode \relax \else
      \ifdim\lastskip<1.5em \hskip-\lastskip
      \hskip1.5em plus0em minus0.5em \fi \nobreak
      \vrule height0.75em width0.5em depth0.25em\fi}

\makeatletter
\renewcommand\section{\@startsection{section}{1}{\z@}%
                                  {-3.5ex \@plus -1ex \@minus -.2ex}%
                                  {2.3ex \@plus.2ex}%
                                  {\normalfont\large\bfseries}}

\renewcommand\subsection{\@startsection{subsection}{1}{\z@}%
                                  {-3.5ex \@plus -1ex \@minus -.2ex}%
                                  {2.3ex \@plus.2ex}%
                                  {\normalfont\normalsize\bfseries}}
\makeatother

\begin{document}

\setlength{\parskip}{5pt} 



\begin{center}

\Large
{\bf Evaluation of a Supervised Learning Approach \\ for Stock Market Operations \\}

\vspace{0.4cm} 

\large

Marcelo S. Lauretto$^1$,  B\'{a}rbara B. C. Silva$^1$ and Pablo M. Andrade$^2$

$^1$EACH$-$USP, $^2$IME$-$USP.
\end{center}

\normalsize


\section{Introduction}

Stock markets play a fundamental role in the countries' economies, 
since they allow companies to raise funds for their investments 
in technology, expansion or infra-structure by selling stocks to the public. 
At the same time, stocks are, for the stockholders, important assets 
that can help to maintain or increase the investor's wealth for future use, like 
retirement, education, etc.
%
On the other hand, stock prices are volatile and depend on several factors like companies' 
performances, economic activity, etc. Hence, investors and funds managers usually must 
constantly monitor the behavior of stock prices, in order to take correct trading 
decisions and to avoid excessive exposition to risky stocks.  

Data mining techniques have been widely proposed for stock market analysis 
in order to identify some patterns in price time series. 
A common premise is that such underlying patterns may be suitably used for 
price forecasting, for operation strategies advices or even for automatic trading. 
In these approaches, usually the attribute vectors 
consist of traditional technical indicators, computed from prices and volumes 
time series.




The objective of this work is to perform an empirical evaluation of 
\textit{Random Forests} for the task of advising trade operations in 
the BM\&F/BOVESPA stock market. We propose a supervised learning 
approach, in which the features are standard technical indicators 
and the classes correspond to three possible actions: 
\textit{Buy-Sell}, \textit{Sell-Buy} or \textit{No action}. 
The evaluation is conducted through a cross validation procedure adapted 
for time series (Hyndman and Athanasopoulos, 2012). 
Three main performance indices are analysed: percentage of opportunities 
seized by the classifier, percentage of successful operations advised by 
the classifier and average return per operation.

\section{Material and Methods}
\vsn

Our case study is based on daily data provided by BM\&F Bovespa Exchange\footnote{\url{http://www.bmfbovespa.com.br/shared/iframe.aspx?idioma=pt-br&url=http://www.bmfbovespa.com.br/pt-br/cotacoes-historicas/FormSeriesHistoricas.asp}}. Raw data is constituted by date, stock identification, prices (opening, minimum, average, maximum, closing), number of trades with the asset and trading value. The study is concentrated on data from January/2010 to October/2012. 

In this preliminary study, we focused on the 68 stocks that integrate the Ibovespa index (BM\&F BOVESPA, 2012), due to their high liquidity and volumes of trading. 

The data processing and tests routines outlined in the next subsections were implemented in the R environment (R Development Core Team, 2011).

\subsection{Random forests}
\vsn

Random forests, introduced by Breiman (2001), are aggregated classifiers composed by 
\textit{ensembles} of trees 
independently induced. The classification of a new instance is made by a voting system, 
where the instance is classified 
by each individual tree and the class ``votes'' are counted. 
Although in most cases the majority criterion is used (the most voted class is assigned), 
it is possible to set up 
lower thresholds such that one class is assigned only if achieves a minimum percentage 
of votes among the trees.   

For the random forest construction, each tree is induced as follows. We denote by $N$ 
the number of examples and by $M$ the number of attributes in the original training set.  

\begin{enumerate}
	\item A bootstrap resample of size $N$ is drawn from the original data, and is used to induce the new tree. 
	\item At each node split, $m \ll M$ attributes are selected at random of the $M$ original attributes, and the 
	best split on these $m$ attributes is used to split the node. The value of $m$ is fixed during the forest construction and may de calibrated by the user. 
	The \textit{randomForest} Package (Liaw and Wiener, 2002), used in this work, sets $m = \sqrt{M}$ 
	as default.
	\item Each tree is grown to the largest extent possible. There is no pruning. 
\end{enumerate}
	
Breiman (2001) shows that the forest error rates increase with the correlation among trees and decrease with the strength of each individual tree in the forest. 
The random sampling of examples and of attributes aim to decrease the trees correlation.


\subsection{Technical indicators}
\vsn

The attribute vectors are constituted by 22 standard technical indicators (Puga et al., 2010), 
computed through the \textit{TTR} Package (Ulrich, 2012):
\begin{itemize}
	\item Simple moving average (SMA) of 3, 13 and 21 days;
	\item Exponential moving average (EMA) of 5, 13 and 21 days;
	\item Rate of change (ROC) of 13 and 21 days;
	\item Stochastic oscillator \%K, slow \%D and fast \%D of 7, 14 and 21 days;
	\item Moving average convergence divergence (MACD) and respective histogram, with short term moving average of 12 days and long term moving average of 26 days;
	\item Relative strength index (RSI) of 9, 14 and 21 days;
\end{itemize}

\subsection{Operation strategies and data classification}
\vsn

A market operation strategy is a predefined set of rules determining an operator's 
action in the market. We consider two operations strategies types, parameterized as follows: $t$ denotes the start day of the strategy, 
$g$ is the maximum expected gain (stop-gain), $l$ is the maximum tolerated loss 
(stop-loss) and $d$ is the maximum duration (in days) of the operation.

\noindent
\textbf{Buy-Sell}($t$, $g$, $l$, $d$): Buy the stock at day $t$ 
  and sell it when the first of the following conditions occurs:
	
	1. Its closing price raises above $g$\% with respect to the price at day $t$;
	
	2. Its closing price falls below $l$\% with respect to the price at day $t$; 
	
	3. After $d$ days, if none of the above cases have occurred in the period 
		      $t+1, t+2, \ldots t+d$.     

\noindent
\textbf{Sell-Buy}($t$, $g$, $l$, $d$): At day $t$, rent a share of the stock, sell it and re-buy an equivalent share of the stock when the first of the following conditions occurs:
	
	1. Its closing price falls below $g$\% with respect to the price at day $t$;
  
  2. Its closing price raises above $l$\% with respect to the price at day $t$; 
	
	3. After $d$ days, if none of the above cases have occurred in the period 
		      $t+1, t+2, \ldots t+d$.     

Notice that in Buy-Sell and Sell-Buy types strategies, return is computed by the 
difference between the sell and buy prices, discounted of the trade costs 
(e.g. brokerage fees). In Sell-Buy strategy, there is an additional rental fee
that must be considered.

An operation strategy is classified as successful if its net return is positive, and 
unsuccessful otherwise. Figure 1 shows two hypothetical examples of applications of 
Buy-Sell strategy. In case (a), the price variation (red line) reaches the expected 
gain ($g$) and the strategy ends successfully (with positive net return) before day 
$t+d$. In case (b), the price variation oscillates between $-l$ and $g$ until the 
day $t+d$, when the strategy ends. Since the net return is negative (the price 
variation is lower than the operation cost), the strategy is unsuccessful.

\begin{figure}[h]
\begin{center}
\includegraphics[width=16cm]{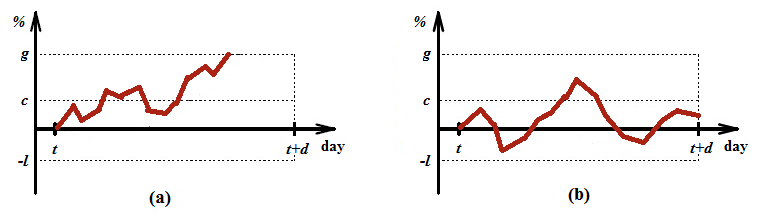}

{\bf Figure 1:} Buy-Sell strategy application examples (adapted from Stern et al, 2008).
\end{center}
\end{figure}  

The dataset classification is performed in the following way. For fixed parameters 
$g$,$l$ and $d$, we verify the success/failure of strategies Buy-Sell($t$, $g$, $l$, $d$)
and Sell-Buy($t$, $g$, $l$, $d$) for each day $t$ in the historical data. If some of these strategies is successful, assign it to day $t$. If none of them is successful, a \textit{No action} class is assigned.  
For convenience, we adopt the following class notation: 
$1=$\textit{Buy-Sell}, $0=$\textit{No action} and $-1=$\textit{Sell-Buy}.



%
%

Notice that there are no \textit{a priori} optimal values for the parameters $g$, $l$ and $d$, since they depend, for example, on the stock price variability, and are strongly dependent each other. 
So, we implemented an automated procedure for setting these parameters, which is described in the next Subsection. 


\subsection{Cross validation}
\label{avalia}

For time series data, the usual $k-$fold or leave-one-out schemes 
are not adequate, due to the high dependency among observations.  
We applied the procedure proposed by Hyndman and Athanasopoulos (2012) (Section 2/5), 
which is similar to the leave-one-out, except that the training set 
consists only of observations that occurred prior to the observation 
that forms the test set. Thus, no future observations can be used
in constructing the classifier. This approach requires that the earliest 
observations are used only for training and are not considered as test sets. 

Denote by $T$ the total length of the dataset, and suppose $k$ observations are required to produce a reliable training. Then the process works as follows. 

\begin{enumerate}
 \item Repeat the following step for $i=1,2,\dots,T-k:$
 \item Build the random forest using the observations at times 
     $i,i+1,i+2,\ldots,i+k-1$, and test it in the observation at time $k+i$. 
     Account the hit/miss (by comparing the predicted and the real classes) 
     and the corresponding return, if any operation strategy has been devised 
     by the forest.  
 \item Compute the total accuracy and net returns obtained for the $T-k$ test samples.
\end{enumerate}

After the above procedure, we obtain a $3 \times 3$ confusion matrix in the form below, were 
rows represent real classes and columns represent predicted classes. The cell $n_{i,j}$ denotes 
the number of test examples of class $i$ that have been classified by the random forests 
as class $j$, for $i, j \in \{-1, 0, 1\}$.

\begin{center}
\begin{tabular}{ccccc} \hline
  & & \multicolumn{3}{c}{Predicted class} \\ 
  & & -1 & 0 & 1 \\ \hline
  \multirow{3}{*}{Real class}  
      & -1 & $n_{-1,-1}$ & $n_{-1,0}$ & $n_{-1,1}$ \\ 
      &  0 & $n_{0,-1}$ & $n_{0,0}$ & $n_{0,1}$ \\ 
      &  1 & $n_{1,-1}$ & $n_{1,0}$ & $n_{1,1}$ \\ \hline
\end{tabular}
\end{center}

\vspace{0.2cm}

Three performance indicators were considered in this work:

\begin{itemize}
	\item $\SeizOport$ : Rate of seized opportunities: ratio between the number of successful 
	operations and the number of opportunities: 
	\[ \SeizOport = \frac{n_{-1,-1} + n_{1,1}} {n_{-1,-1} + n_{-1,0} + n_{-1,1} + n_{1,-1} + n_{1,0} + n_{1,1}} \]

	\item $\SuccOper$ : Rate of successful operations: ratio between the number of successful operations and the total number of devised operations:
	\[ \SuccOper = \frac{n_{-1,-1} + n_{1,1}} {n_{-1,-1} + n_{0,-1} + n_{1,-1} + n_{-1,1} + n_{0,1} + n_{1,1}} \]
	
	\item $\AvgRetOper$: Average return per operation: ratio between the sum of net returns 
	yielded by the devised strategies (disregarding success or failure) and the total number of 
	devised operations. 
\end{itemize}

These performance indicators are combined in a single \score, defined by the 
following convex combination:
\[
	\Score = 0.10 \ \SeizOport + 0.85 \ \SuccOper + 0.05 \ \AvgRetOper
\]

These weights were set in order to turn the score a conservative function, 
in the sense that it favors strategies with high rates of successful operations, even though achieving lower values in the other indicators.

The procedure for setting the values of parameters $g$, $l$ and $d$ is as follows.  
First, we define, for each parameter, a set of candidate values. In the present
study, these sets are:
\begin{itemize}
	\item $g \in \{10\%, 15\%, 20\% ... 35\%\}$ 
	\item $l \in \{3\%, 6\%, 9\%, ... 15\%\} $
	\item $d \in \{10, 15, 20, ... 35\} $.
\end{itemize}

For each value in the grid above, the operations strategies are simulated on 
the data series and the examples are labeled with the corresponding classes. 
The cross validation is run and the indicators $\SeizOport$, $\SuccOper$, 
$\AvgRetOper$ and $\Score$ are computed. For each stock, we choose the 
values of $g$, $l$, $d$ that maximize the function $\Score$. 

In our simulations, the operation cost is assumed as $c = 1\%$, and the stock 
rental fee is assumed as $0.05\%$ per day.  

For setting the operation strategies parameters, the cross validation 
procedure uses data of 2010 for training and data of 2011 
for testing. After the parameters setup, a new cross 
validation is run for a final evaluation of the optimal parameters, 
taking data of 2011 for training and data of 2012 for testing. 
  
\section{Results and Conclusions}

Figure 2 presents the indicators $\SeizOport$, $\SuccOper$, $\AvgRetOper$ 
and $\Score$ for the 30 stocks with greater score values computed 
on 2012 data test. 
The proposed method yields more than $80\%$ of successful 
devised operations for almost all stocks, and also yields more 
than $70\%$ of seized opportunities for 22 of the 30 stocks ($73\%$). 
The average returns per operation are also expressive, achieving 
for the majority of stocks $4\%$ or more. These returns may be 
considered high, since one strategy operation lasts at most 
$35$ trading days (see previous Section). 
   
\begin{figure}[h]
\begin{center}
\includegraphics[width=16cm]{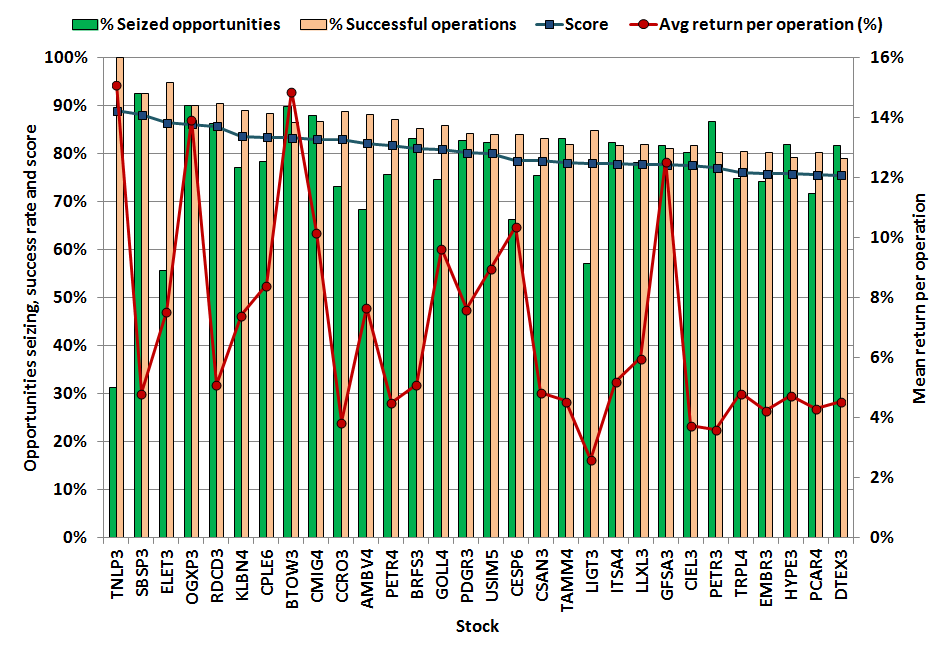}

Figure 2: Performance indicators $\SeizOport$, $\SuccOper$, $\AvgRetOper$ and 
        $\Score$ for the 30 stocks with maximum score values.
\end{center}
\end{figure}

The preliminary results presented in this work are very promising and motivate
several extensions. Some examples are the introduction of other performance 
indices; the inclusion of other technical indicators; performance analyses carried independently for Buy-Sell and Sell-Buy strategies; the 
incorporation of more than one parameter by each strategy type; comparison 
of the performance with other classification algorithms; the introduction 
of slippage in the model, and sereral others.

The authors are grateful for the support of EACH-USP and IME-USP, 
to the Coordenação de Aperfeiçoamento de Pessoal de Nível 
Superior (CAPES), Conselho Nacional de Desenvolvimento 
Científico e Tecnológico (CNPq) and Fundação de Apoio à 
Pesquisa do Estado de São Paulo (FAPESP).

\section*{References}

\noindent
S. Arlot and A. Celisse (2010). A survey of cross-validation procedures for model selection. \textit{Statistical Surveys} 4, 40$-$79.

\noindent
L. Breiman and A. Cutler (2012). \textit{Random Forests}. Available at \url{http://www.stat.berkeley.edu/~breiman/RandomForests/cc_home.htm}. 

\noindent
BM\&F BOVESPA. \textit{Índice Bovespa - Ibovespa}. Available at \url{http://www.bmfbovespa.} 
 \url{com.br/indices/ResumoCarteiraTeorica.aspx?Indice=IBOVESPA\&idioma=pt-br}.  

\noindent
R. J. Hyndman and G. Athanasopoulos (2012). Forecasting: principles and practice. Online textbook available at \url{http://otexts.com/fpp/}.

\noindent
A. Liaw and M. Wiener (2002). Classification and Regression by randomForest. \textit{R News} 2(3), 18$-$22.

\noindent
M. Miró-Julià, G. Fil-roig and A. P. Isern-deyà (2010). 
  Decision Trees in Stock Market Analysis: Construction and Validation. 
  In: N. García-Pedrajas et al. (Eds.): \textit{IEA/AIE 2010}, Part I, LNAI 6096, p. 185$-$194.

\noindent 
NYSE Euronext (2012). \textit{Why We Invest}. Available at \url{https://nyse.nyx.com/} \\ 
\url{financial-literacy/all-about-investing/investing-basics/why-we-invest}.

\noindent
R. Puga, M. Rodrigues, G.Cerbassi (coord) (2010). \textit{Formação de traders: faça dinheiro na bolsa com a análise técnica}. Rio de Janeiro: Campus.

\noindent
J. R. Quinlan (1986).  Induction of decisions trees. \textit{Machine Learning} 1, 81-106.

\noindent
J. M. Stern, F. Nakano, M. S. Lauretto, C. O. Ribeiro (1998). Algoritmo de Aprendizagem para Atributos Reais e Estratégias de Operação em Mercado. In: \textit{Sixth Iberoamerican Conference on Artificial Intelligence - IBERAMIA'98}, Lisboa.

\noindent
R Development Core Team (2011). \textit{R: A Language and Environment for Statistical
  Computing}. R Foundation for Statistical Computing, Vienna, Austria. \\
  \url{http://www.R-project.org}      

\noindent
J. Ulrich (2012). \textit{The TTR Package Reference Manual}. Available at \\ \url{http://cran.r-project.org/web/packages/TTR/index.html} 

\end {document}